\documentclass{article}
\usepackage{spconf,amsmath,graphicx}
\usepackage{multirow}
\usepackage{booktabs} 
\usepackage{subcaption}
\usepackage{multirow}
\let\OLDthebibliography\thebibliography
\renewcommand\thebibliography[1]{
    \OLDthebibliography{#1}
    \setlength{\parskip}{0pt}
    \setlength{\itemsep}{0pt plus 0.3ex}}
\usepackage{amsmath}

\newcommand\correspondingauthor{\thanks{$^{*}$Corresponding author}}

\title{Improved Meta Learning for Low Resource Speech Recognition}
\name{Satwinder Singh \qquad Ruili Wang \qquad Feng Hou$^*$\correspondingauthor}
\address{School of Mathematical and Computational Sciences, Massey University, Auckland, New Zealand}
\begin{document}
%
\maketitle
\begin{abstract}
We propose a new meta learning based framework for low resource speech recognition that improves the previous model agnostic meta learning (MAML) approach. The MAML is a simple yet powerful meta learning approach.  However, the MAML presents some core deficiencies such as training instabilities and slower convergence speed. To address these issues, we adopt multi-step loss (MSL). The MSL aims to calculate losses at every step of the inner loop of MAML and then combines them with a weighted importance vector. The importance vector ensures that the loss at the last step has more importance than the previous steps. Our empirical evaluation shows that  MSL significantly improves the stability of the training procedure and it thus also improves the accuracy of the overall system. Our proposed system outperforms MAML based low resource ASR system on various languages in terms of character error rates and stable training behavior.
\end{abstract}
\section{Introduction}
\label{sec:intro}
Modern deep learning based end-to-end (E2E) models have lately become extremely popular in the speech community \cite{singh2021deepf0} and have achieved a significant milestone in terms of performance. These systems have been deployed under commercial domains as they have shown consistently lower word error rates that are close to 1-2\% \cite{baevski2020wav2vec}. 
The modern ASR systems are mostly trained in end-to-end (E2E) fashion without requiring resources like a pronunciation dictionary and a language model as separate modules. These systems are able to achieve such a high degree of accuracy mainly because they are trained on various high performance large vocabulary datasets. However, these E2E systems tend to perform much worse for the languages that do not have such large quantities of annotated data.   
\par
Among roughly 7000 languages spoken across the world, there are only around 100 languages that have well-established speech recognition systems \cite{chuangsuwanich2016multilingual}. The rest of the languages are considered as low resource languages because they do not have a huge amount of annotated speech data, strong pronunciation dictionaries, and a huge collection of unpaired texts. A lot of progress has been made in low resource speech recognition, which includes efforts like transfer learning \cite{tu2019end} and multilingual training \cite{zhou2018multilingual}. Recently, a new paradigm, meta learning has been explored for low resource speech recognition  \cite{hsu2020meta}. Meta learning (also known as learning to learn) is a machine learning technique, where learning is done on two levels. On one level  (inner loop) model acquires task specific knowledge, whereas the second level (outer loop) facilitates task across learning \cite{antoniou2018train}.  
\par
Previously, Hsu et al. \cite{hsu2020meta} proposed a meta learning framework based on the MAML approach for ASR for low resource language. The proposed framework outperformed the no-pretraining and multi-lingual training settings. Similarly, Winata et al. \cite{winata2020learning} incorporated the MAML approach for the few shot accent adaptation task for the English. The MAML approach in general is a very straightforward and powerful approach. However, it is prone to numerous problems, including unstable training and slow convergence speeds. These issues also impact the generalizability of the model. Thus, to deal with these issues, in this paper we adopt the multi-step loss \cite{antoniou2018train}, which is introduced to stabilize the meta training procedure. The meta training approach with multi-step loss calculates the inner loss after every inner step updates and later computes the weighted sum of all the inner losses.  
\par
We evaluated our proposed approach on 10 different languages present in the Common Voice v7.0 dataset. All these languages are represented in form of a low resource setting where the language data ranges from 0.5 hours to 300 hours. We find that our approach indeed improves the training instabilities of the MAML approach, which in turn improves the overall accuracy of the model. 

 \section{Related work}
\label{sec:1}
\subsection{Meta Learning}
\label{sec:2}
Meta learning is not a new idea but begin to gain attention in recent times. Recently, in the context of deep learning, meta learning comes into the limelight due to its wide range of applications and advantages. Meta learning helps to generalize to various tasks faster with few steps and examples. Literature suggests the application of meta learning in two ways where the first is learning a better initialization of network parameters  \cite{mishra2017meta} and the second is learning a strategy or procedure for updating the parameters of the network \cite{finn2017model}, \cite{snell2017prototypical}. 
\par
Meta learning has been applied to a range of research domains including various computer vision tasks, natural language processing and recently automatic speech recognition. In the computer vision area, meta learning has been exploited for the few-shot image classification task \cite{ravi2017OptimizationAA}, object detection \cite{phaphuangwittayakul2021fast} and  video generation \cite{wang2019few}. In the natural language processing domain, meta learning has shown promising results in neural machine translation (NMT) for resource constraint languages \cite{gu2018meta}. Apart from this, recently researchers have tried meta learning for speech processing tasks, such as automatic speech recognition \cite{hsu2020meta}, speaker adaptation \cite{klejch2018learning}, \cite{klejch2019speaker} and recognition \cite{kye2020meta}, cross-lingual \cite{hou2021meta} and cross-accent adaption \cite{winata2020learning}. 

\subsection{Low Resource Speech Recognition}
The development of a speech recognition system for a low resource languages has been a very active research area for the past few years. The regular E2E ASR systems designed for resource rich languages seem not to work for low resource languages due to the lack of annotated speech data or other resources. There have been many attempts made to alleviate the scarcity of labelled speech data. These efforts include, speech data augmentation \cite{park2019specaugment}, transfer learning \cite{tu2019end}, multilingual  \cite{zhou2018multilingual}, cross-lingual \cite{hou2021meta} and multi-task learning \cite{hsu2020meta}. Recently, unsupervised cross-lingual wave2vec 2.0 XLSR model \cite{conneau2020unsupervised} shown a huge performance boost compared to other previous state-of-the-art models. Further, there have been recent attempts to explore a new research direction of meta learning for low resource languages. The idea is to extract meta parameters learned over multiple source languages and then bootstrap these learned meta parameters to fine-tune on the target languages. The whole process can be seen as learning a model that can perform fast adaptation to target languages with few epochs and data samples. As fine-tuning requires few training samples, this process of meta learning is totally aligned with our proposed framework of ASR for low resource languages.   

\section{Proposed System}
Our proposed system consists of two core components. The first is an ASR model that acquires language specific knowledge and the second is a multi-step loss based model agnostic meta learning algorithm. 

\subsection{The ASR Model}
\begin{figure}[!t] 
  \centering
  \includegraphics[width=7cm]{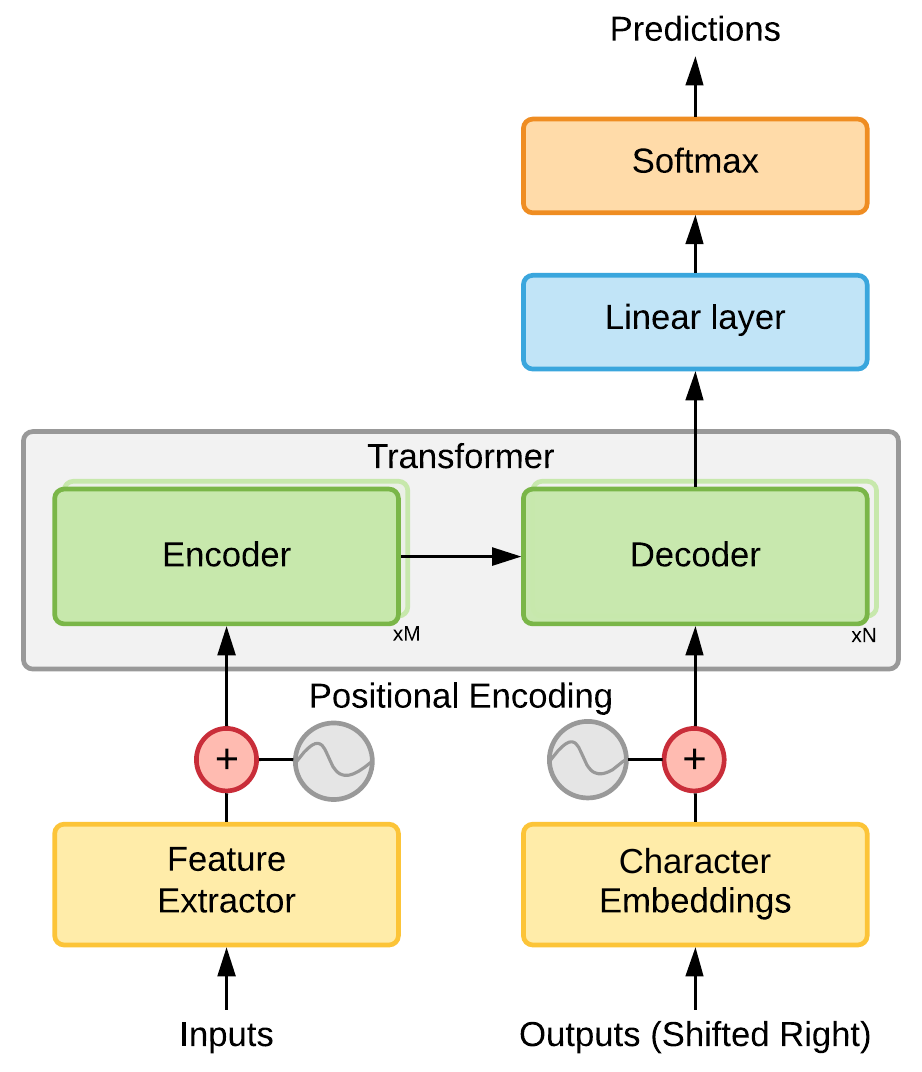} 
  \caption{The Transformer model for ASR }
  \label{fig:mamlvsmaml++}
\end{figure}
\begin{figure}[!h] 
  \centering
  \includegraphics[width=9cm]{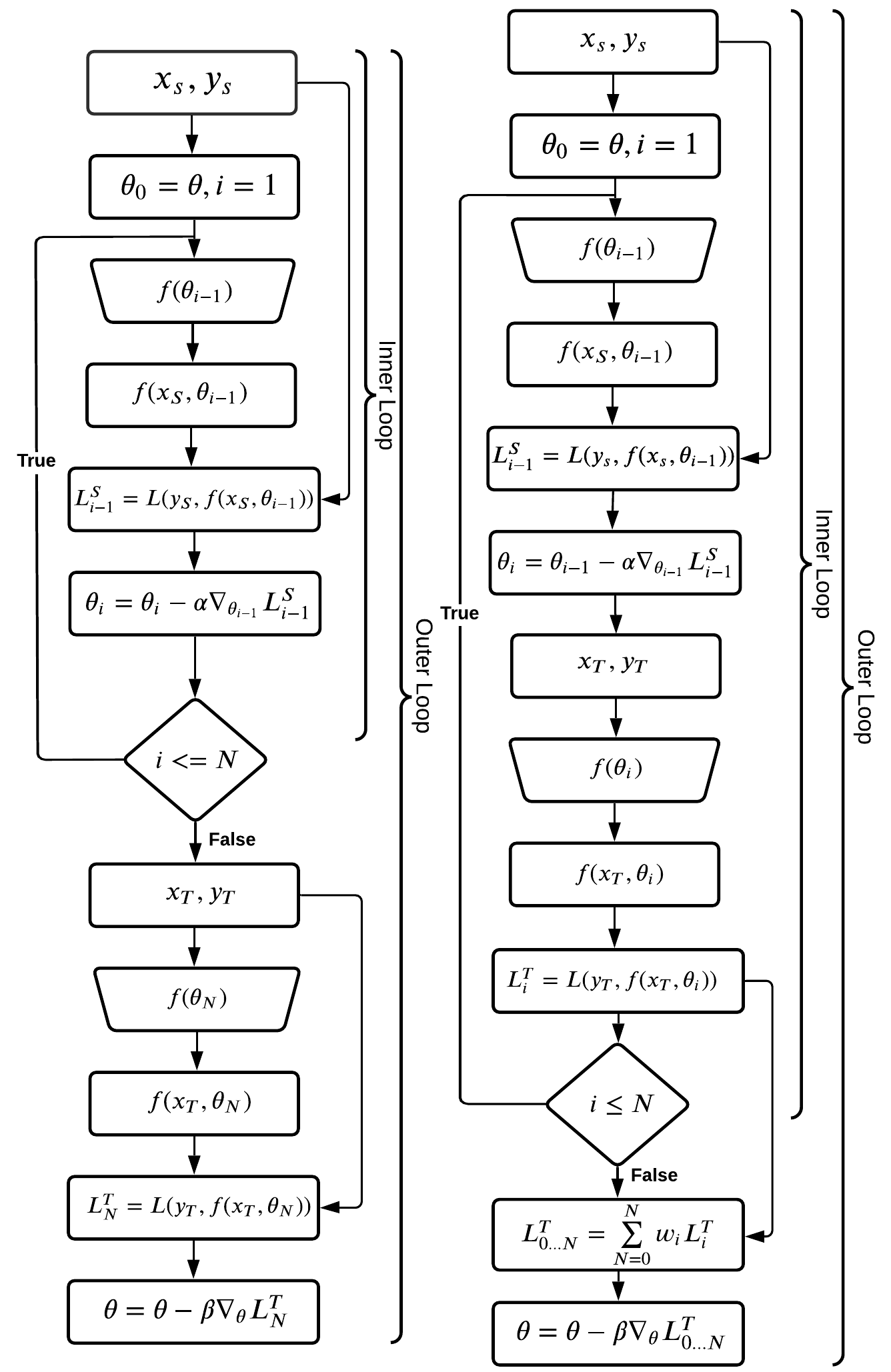} 
  \caption{MAML (Left) vs MAML with MSL (Right) (adopted from \cite{antoniou2018train})}
  \label{fig:mamlvsmaml++}
  
\end{figure}

For our proposed system, we adopt the transformer ASR model \cite{vaswani2017attention} as our language specific model. The transformer model is a sequence-to-sequence model based on the encoder-decoder architecture. The proposed model extracts the input features using the learnable VGG based convolutional neural network (CNN) model \cite{simonyan2015very}. The input embeddings produced by the feature extractor are then fed to the encoder module through the positional encoding setup. The positional encoding setup generates a vector that is served as context for the symbols. Afterward, the outputs of the encoder module are passed on to the decoder module, where a multi-head attention mechanism is employed on these encoder outputs. The attention mechanism applies masking in the decoder block to restrict the attention layer from attending to any future tokens. Finally, the output of the decoder block goes through a linear and softmax layer and generates the predictions. The entire training process is optimized by maximizing the log probability using next step prediction based on the last output token. In the following equation,  $x$, $y_i$ and $y'_{i-1}$ are the input character, next predicted character, and true label of the last character, respectively. 
\begin{equation}
   \max_\theta \sum \limits_i \log P (y_i|x,y'_{i-1};\theta)
\end{equation}

\subsection{Meta Learning Setup}
In general, our proposed meta learning setup aims at learning the initial parameters for the model in a way that it can be quickly adapted to new languages with a fewer number of gradient descent steps. We adopt multi-step loss from MAML++ \cite{antoniou2018train} procedure over standard MAML as MAML tends to have unstable training procedure. This can affect the overall speed of convergence, and also has a negative impact on the accuracy of the model. Figure \ref{fig:mamlvsmaml++} shows the computational graph for both MAML and MAML with multi-step loss. We select support samples $(x_S, y_S)$ and validation data samples $(x_T, y_T)$ from our source languages set. We start optimizing the inner loop by initializing our ASR model $f$ with  $\theta_0=\theta$. Afterwards, the ASR model produces logits $f(x_S, \theta_{i-1})$ by using samples from training set and parameters $\theta_{i-1}$. Here $i$ represents $i^{th}$ step of total $N$ steps. In the next step, loss $L^S_{i-1}$ is computed over true labels $y_S$ and logits. Further, the $L^S_{i-1}$ is utilized to update the current parameters of the model.   
\par
The inner optimization loop of our MSL MAML approach differs from MAML, where instead of using $\theta_N$ parameters for computing target set loss, our MSL MAML approach goes on using $\theta_i$ parameters. After completing the inner loop, we obtain $N$ target set losses as in Eq. \ref{eq:2}, which can be seen as a multi-step loss, where $w_i$ is the weight of step $i$ and specify the importance of per step target loss. Initially, all the losses have approximately the same importance, while later in training more importance is given to the losses on later steps. This way the model gradually steps towards the MAML loss, ensuring there is no issue of gradient degradation. Finally, these losses are combined together using a weighted sum of per step losses. The combined weighted loss is then used to update the outer loop parameters $\theta$. The advantage of calculating per step loss is reducing the gradient vanishing and exploding problem of the original MAML. Following \cite{winata2020learning} and \cite{hsu2020meta}, we only compute first order approximation of $\theta$. 
\begin{equation}
\label{eq:2}
    L^T_{0...N}=\sum \limits_{N=0}^{N}w_iL^T_i
\end{equation}

\section{experimental Setup}
\subsection{Dataset}
For our experiments, we choose Common Voice dataset version 7 \cite{ardila2020common}. The data in Common Voice is a crowdsourced public dataset and contains many languages including resource rich and low resource languages. We select 10 low resource languages and the description is represented in Table \ref{table:data}. Some of the languages are very low resource having just a few hours of data. The audio from all the languages is downsampled to 16 kHz and labels are preprocessed to remove any kind of special symbols.  

\begin{table}[!h]
\centering
\caption{The selected low resource languages from the Common Voice dataset v7.0 and the total amount of speech data in terms of hours.}
\label{table:data}
\begin{tabular}{c|c|c}
\bottomrule
\textbf{ID}         & \textbf{Languages}      & \textbf{Hours}  \\ \toprule
ar         & Arabic         & 85     \\ \hline
as         & Assames        & 1      \\ \hline
hi         & Hindi          & 8      \\ \hline
lt         & Lithuanian     & 16     \\ \hline
mn         & Mongolian      & 12     \\ \hline
or         & Odia           & 0.94   \\ \hline
fa         & Persian        & 293    \\ \hline
pa-IN      & Punjabi        & 1      \\ \hline
ta         & Tamil          & 198    \\ \hline
ur         & Urdu           & 0.59   \\ \hline
\multicolumn{2}{c|}{Total} & 615.53 \\ \toprule
\end{tabular}
\end{table}

\begin{table*}[!t]
\centering
\caption{The average experimental results in terms of character error rate (CER in \%) on 5 target languages. We have not fine-tune our model on the languages that are present in the pretrain source language sets. These cells are represented by hyphen (-)}
\label{table:result}
\begin{tabular}{c|c|c|c|c|c|c|c|c|c|c}
\bottomrule
\multirow{2}{*}{Pretrain languages} & \multicolumn{10}{c}{Finetune}                                                                                                                                \\ \cline{2-11} 
                                    & \multicolumn{2}{c|}{Hindi} & \multicolumn{2}{c|}{Mongolian} & \multicolumn{2}{c|}{Persian} & \multicolumn{2}{c|}{Arabic}    & \multicolumn{2}{c}{Tamil}      \\ \toprule
                                    & MAML    & Our              & MAML      & Our                & MAML     & Our               & MAML          & Our            & MAML           & Our            \\ \hline
{[}fa, ar, ta{]}                    & 70.51   & \textbf{70.47}   & 61.05     & \textbf{60.52}     & -        & -                 & -             & -              & -              & -              \\ \hline
{[}ar, mn, lt{]}                      & 71.61   & \textbf{71.37}   & -         & -                  & 47.96    & \textbf{45.45}    & -             & -              & 40.96          & \textbf{35.17} \\ \hline
{[}or, pa-IN, hi, ur, as{]}         & -       & -                & 62.26     & \textbf{59.50}      & 52.42    & \textbf{52.41}    & \textbf{36.00} & 36.09 & \textbf{45.96} & 46.60          \\ \toprule
\end{tabular}
\end{table*}                                                     

\subsection{Methodology}
Our model receives spectrogram as an input. These spectrogram inputs then go through a VGG based 6-layered CNN feature extractor. We use 2 encoder layers and 4 decoder layers with 8 multi-head attention layers. Our model produces input and output of dimension 512, whereas the inner layer has 2048 dimensions. We set the dropout value to 0.1 and keys and values dimensions to 64. We multilingually pretrain our model for 100K iterations on the source language set. We put together 3 source language  sets where one set includes \textbf{fa, ar} and, \textbf{ta}. The other set has \textbf{ar, mn} and, \textbf{lt} and the last set consists of \textbf{or, pa-IN, hi, ur,} and \textbf{as}. During the fine-tuning phase, we fine-tune the model on our target languages (\textbf{hi, mn, fa, ar} and \textbf{ta}) one by one for 10 epochs. The model is then evaluated on a test set of target language using beam search with a beam size of 5. 

\begin{figure}[!b] 
  \centering
  \includegraphics[width=0.5\textwidth]{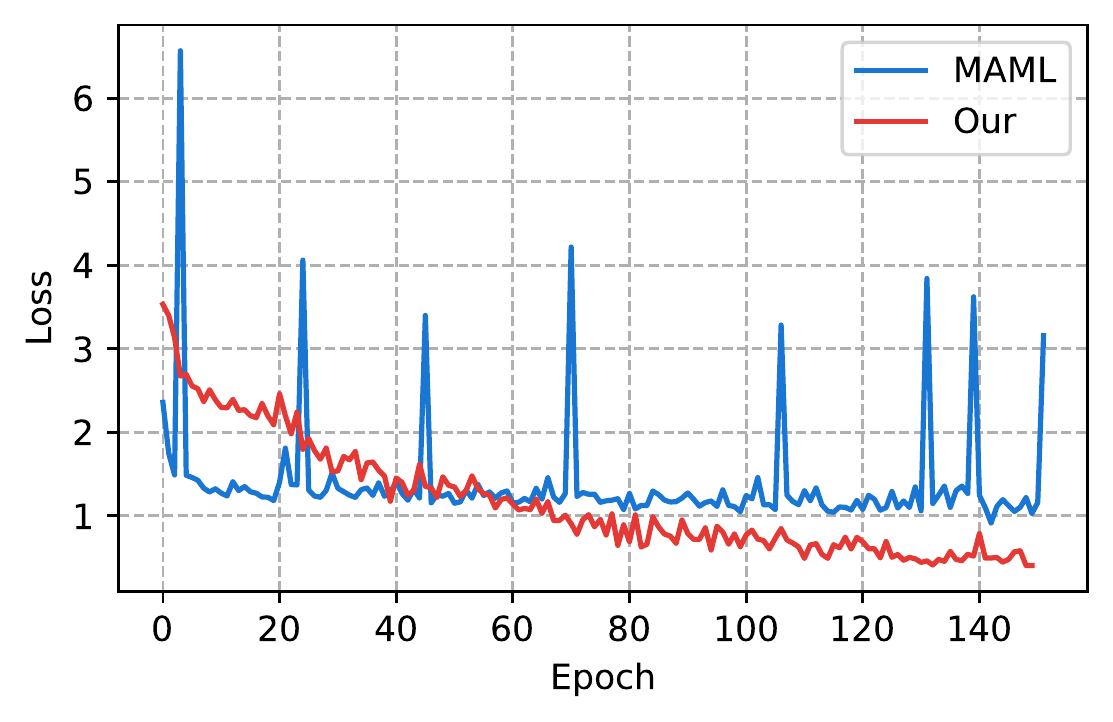} 
  \caption{Training curve of MAML vs our approach. The training loss curve for MAML shows unstable peaks whereas our approach shows more consistent loss curve.}
  \label{fig:curve}
\end{figure}

\section{Results and discussion }      
\subsection{Model's Accuracy Analysis}
We evaluate the performance of our proposed MSL MAML approach on 10 languages from the Common Voice dataset. Our proposed approach showcases consistent improvement in character error rates (CER in \%)  over the standard MAML approach. The detailed results are presented in Table \ref{table:result}.
On source languages set \textbf{[fa, ar, ta]} our approach achieves 70.47\% and 60.52\% of CER on Hindi and Mongolian language,  respectively. Our proposed model shows around 1\% of improvement over standard MAML on the Mongolian language. On set \textbf{[ar, mn, lt]} our approach slightly performs better than MAML on the Hindi language. On the same set, our approach outperforms the current MAML approach with 5.23\% and 14.13\% of relative improvement on Persian and Tamil language, respectively. 
\par
Further, the Mongolian language demonstrates 4.43\% of relative improvement over MAML on set \textbf{[or, pa-IN, hi, ur, as]}. Mostly, on this pretrain language set both MAML and our approach report similar results on Persian and Arabic languages. Interestingly, the MAML marginally outperforms our approach on the Tamil language. Overall, our approach shows consistent improvements across all the pretrain sets, where excellent performance is observed on \textbf{[ar, mn, lt]}.  

\subsection{Training Performance Analysis}

The multi-step loss indeed stabilizes the training process of MAML as shown in Figure \ref{fig:curve}. The primary driver of instability in MAML is the gradient degradation problem while training deep network \cite{antoniou2018train}. Our approach resolved this issue using multi-step loss, where the model is evaluated at each step against its validation set. Further, importance weight also makes sure later step loss has more importance. It also improves the convergence speed of the model as shown in Figure \ref{fig:curve}. 
\vspace{-7mm}
\section{Conclusions}
In this paper, we propose a multi-step loss based meta learning approach for speech recognition for low resource languages. The proposed system improves the inner loop optimization for the MAML algorithm, which results in a more stabilized training procedure. Our empirical results show that multi-step loss indeed improves the overall training procedure and also has a positive impact on the accuracy of the model. Apart from this, our model also trains faster as compared to MAML. 
In the future, we plan to conduct more experiments with more low resource languages. We would extend our experiments with different combinations of languages on the basis of their phonetic structures, geographic areas, and language family.
\vspace{-6mm}
\section{Acknowledgement}
This work is supported by the 2020 Catalyst: Strategic New Zealand - Singapore Data Science Research Programme Fund by  Ministry of Business, Innovation and Employment (MBIE), New Zealand.

\bibliographystyle{IEEEbib}
\bibliography{strings,refs}

\end{document}